\documentclass[letterpaper]{article} % DO NOT CHANGE THIS
\usepackage[submission]{aaai23}  % DO NOT CHANGE THIS
\usepackage{times}  % DO NOT CHANGE THIS
\usepackage{helvet}  % DO NOT CHANGE THIS
\usepackage{courier}  % DO NOT CHANGE THIS
\usepackage[hyphens]{url}  % DO NOT CHANGE THIS
\usepackage{graphicx} % DO NOT CHANGE THIS

\usepackage{amsmath}
\usepackage{amssymb}
\usepackage{subfigure}
\usepackage{color}
\usepackage{multirow}

\usepackage{natbib}  % DO NOT CHANGE THIS AND DO NOT ADD ANY OPTIONS TO IT
\usepackage{caption} % DO NOT CHANGE THIS AND DO NOT ADD ANY OPTIONS TO IT
\usepackage{algorithm}
\usepackage{algorithmic}

\usepackage{newfloat}
\usepackage{listings}
\DeclareCaptionStyle{ruled}{labelfont=normalfont,labelsep=colon,strut=off} % DO NOT CHANGE THIS
\floatstyle{ruled}
\newfloat{listing}{tb}{lst}{}
\floatname{listing}{Listing}
\title{Rethinking Blur Synthesis for Deep Real-World Image Deblurring}
\author{
   Hao Wei,
   Chenyang Ge, 
   Xin Qiao, 
   Pengchao Deng
}
\affiliations{
	Xi'an Jiaotong University\\
}

\usepackage{bibentry}
\begin{document}
	
\maketitle

\begin{abstract}
%AAAI creates proceedings, working notes, and technical reports directly from electronic source furnished by the authors. To ensure that all papers in the publication have a uniform appearance, authors must adhere to the following instructions.
% 
In this paper, we examine the problem of real-world image deblurring and take into account two key factors for improving the performance of the deep image deblurring model, namely, training data synthesis and network architecture design. Deblurring models trained on existing synthetic datasets perform poorly on real blurry images due to domain shift. To reduce the domain gap between synthetic and real domains, we propose a novel realistic blur synthesis pipeline to simulate the camera imaging process. As a result of our proposed synthesis method, existing deblurring models could be made more robust to handle real-world blur. Furthermore, we develop an effective deblurring model that captures non-local dependencies and local context in the feature domain simultaneously. Specifically, we introduce the multi-path transformer module to UNet architecture for enriched multi-scale features learning. A comprehensive experiment on three real-world datasets shows that the proposed deblurring model performs better than state-of-the-art methods. 
\end{abstract}

\section{Introduction}
During the recording process, motion blur is often caused by moving objects or shaking devices, which results in poor image quality. Single image deblurring, which aims to restore the sharp content of a blurry image, is a classic problem in computer vision and image processing. Nevertheless, this is a challenging research topic due to its highly ill-posed nature.

\begin{figure}[!t]
	\captionsetup[figure]{name={Figure}}
	\centering
	\subfigure[Blurry input]{
		\label{ITC:Blurry input}
		\includegraphics[scale=0.7]{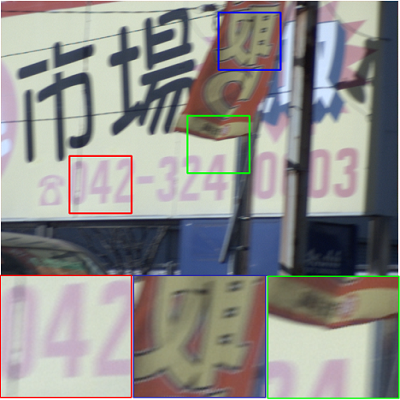}
	}
	\subfigure[MIMO-UNet (GoPro)]{
		\label{ITC:GoPro}
		\includegraphics[scale=0.7]{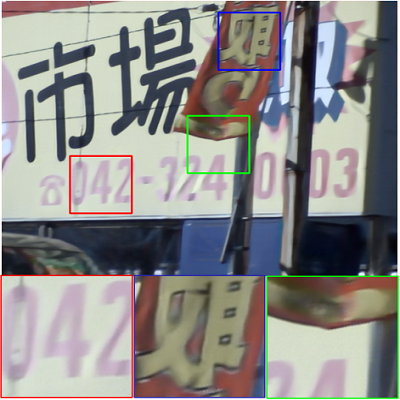}
	}
	\subfigure[MIMO-UNet (REDS)]{
		\label{ITC:REDS}
		\includegraphics[scale=0.7]{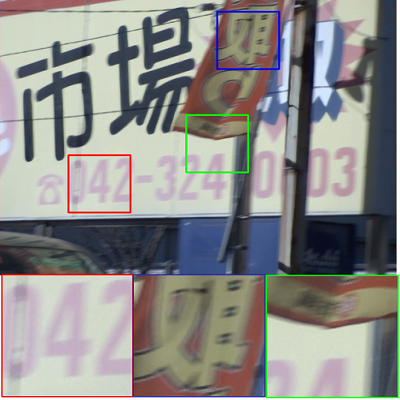}
	}
	\subfigure[MIMO-UNet (Ours)]{
		\label{ITC:Ours}
		\includegraphics[scale=0.7]{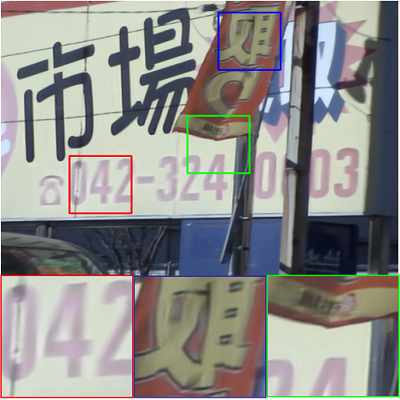}
	}
	\caption{We demonstrate the effectiveness of our proposed blur synthesis method, which is more consistent with how real-world blur is generated. Given a real-world blurry image, the restoration model MIMO-UNet~\cite{MIMOUNet} trained with different training datasets can predict diverse visual deblurred results. In qualitative comparisons, it becomes evident that a realistic blur synthesis method is necessary, as it directly affects the performance of the deblurring model in real-world situations.}
	\label{intro_TC}
\end{figure}
Recently, a number of learning-based approaches~\cite{GoPro, SRN, DMPHN, MPRNet, MIMOUNet, Restormer, NAFNet} have been proposed and almost dominate this field. 
These methods directly learn a mapping function between blurry images and their sharp counterparts in an end-to-end manner. However, these methods rely heavily on synthetic datasets (e.g., GoPro~\cite{GoPro} and REDS~\cite{REDS}) and are not generalizable well to real-world blur. 
To reduce the gap between synthetic and real-world blur, some researchers~\cite{RealBlur, BSD} build the real-world blur datasets using a dual-camera system. However, the acquisition process is time-consuming and labor-intensive and the captured images require sophisticated postprocessing, such as geometric alignment. Furthermore, the collected real datasets are biased towards the specific recording camera used. This directly limits the effectiveness of deblurring models when applied to real-world blurs captured with other hand-held devices. 
Therefore, the limitations above motivate us to rethink the existing blur synthesis methods and develop a realistic data synthesis pipeline that is more consistent with the real-world blur generation process. 

On the other hand, not only the training data but also the network architecture design affects the performance of deep deblurring models for real blur removal. In~\cite{GoPro, SRN, PSS-NSC}, the researchers develop a multi-scale network for progressively deblurring images. Furthermore,~\cite{MIMOUNet} further proposes a multi-input multi-output design using a single UNet~\cite{UNet}. Moreover, some researchers~\cite{DMPHN, Suin, MPRNet} introduce multi-patch and multi-stage schemes in their network architecture design to boost deblurring performance. 
In the above-mentioned methods, the main success comes from the powerful learning ability of deep convolutional neural networks (CNNs) for capturing generalizable image priors. However, CNNs lack the ability to capture long-range pixel dependencies. In order to address this drawback, transformer-based methods~\cite{transformer, MPViT} have been developed for low-level tasks~\cite{IPT, SwinIR, EDT}, for example, image deblurring~\cite{Restormer, Uformer}. However, these transformer-based methods fail to obtain the multi-scale feature representations at the same feature level that are vital to image deblurring~\cite{GoPro}. 

To improve the real-world image deblurring, we consider two perspectives, namely, training data and network architecture. 
For the training data, motivated by the fact that deblurring models trained with existing synthetic datasets (e.g., GoPro and REDS) are not effective for real-world blur removal, we propose a novel realistic blur synthesis pipeline for generating realistic data. Specifically, we firstly employ frame interpolation technology~\cite{FILM} to increase the frame rate of GoPro datasets. The video frames are then converted to sharp RAW images via reverse learnable image processing (ISP) pipeline~\cite{CycleISP} and averaged to obtain blurry RAW images. Note that the blurs are synthesized in the RAW domain, which corresponds with the process by which real blurs are generated when the hand-held devices accumulate signals during exposure. Lastly, we employ forward learnable ISP to reconstruct the realistic blurry RGB images based on blurry RAW data. The learnable ISP is not restricted to a particular camera and can be easily generalized to other hand-held devices. As shown in Figure~\ref{intro_TC}, the deblurring model trained with our synthetic data shows favorable results when compared to the model trained with previous synthetic datasets. 

For the design of the network achitecture, we propose a multi-path transformer-based UNet (MPTUNet), which is capable of capturing long-range and local dependencies simulatneously. Specifically, we introduce a multi-path transformer module (MPTM) as the basic component of the UNet architecture. The MPTM performs overlapping convolutional patch embedding to extract the patches with different scales and then the patches are independently fed into the transformer block in the parallel manner. Moreover, we plug the residual block into MPTM to enable local modelling capabilities. As a result of the aforementioned designs, the MPTUNet is capable of achieving superior or comparable performance with fewer parameters compared to the state-of-the-art deblurring models.

The main contributions are summarized as follows:
\begin{itemize}
	\item We propose a noval realistic blur data synthesis pipeline. The pipeline aims to improve the generalization ability of deblurring models for real blur removal.
	\item To take advantage of the local modeling ability of CNNs and non-local modeling ability of transformer, we develop a deblurring network by introducing multi-path transformer module into UNet. The multi-path transformer module learns enriched multi-scale feature representations that are useful for image deblurring.
	\item We quantitatively and qualitatively evaluate the deblurring models trained with our synthesis data on the real-world datasets and verify the effectiveness of the proposed realistc blur synthesis pipeline.
	\item The proposed deblurring network performs well on real-world datasets quantitatively and qualitatively against state-of-the-art algorithms.
\end{itemize}

\section{Related work}
\subsection{Image Deblurring Datasets}
Datasets play an important role in the development of image deblurring algorithms~\cite{Kohler, Lai, Levin, Shen, HIDE, DVD, GoPro, REDS, RealBlur, BSD, 4KRD}. Traditionly, blurry images have been simulated by convolving sharp images with uniform or non-uniform blur kernels~\cite{Levin, Kohler}. In order to evaluate the performance of blind image deblurring algorithms, Lai et al.~\cite{Lai} collect two datasets that contain real blurry images and synthetic blurry images. It is however difficult to apply deep learning-based deblurring methods to these datasets due to their limited scale. 
To remedy this problem, Nah et al.~\cite{GoPro} employ kernel-free method for large-scale dataset generation. Inspired by ~\cite{GoPro}, Su et al.~\cite{DVD} release a video deblurring dataset consisting of 71 video sequences. Recently, Deng et al.~\cite{4KRD} create a new dataset to handle blurs in ultra-high-definition videos. However, when employ the low-frame-rate videos to generate blurry images, the unnatural artifacts will appear due to large motion. To avoid this, Nah et al.~\cite{REDS} introduce video frame interpolation to generate the high-frame-rate videos and collect realistic and diverse scenes dataset. 

Nevertheless, the existing synthetic datasets cannot generalize well to real-world blurred images as will be shown in our experiments. Therefore, several researchers begin to design complex image acquisition systems to capture pairs of blurred and sharp images. Rim et al.~\cite{RealBlur} collect un-aligned real-world datasets under low-light conditions using a dual-camera system. Zhong et al.~\cite{BSD} use a beam splitter system with two synchronized cameras to capture paired blurry and sharp images. Although the fact that the above two datasets are real-captured, they require either sophisticated alignment procedures or lengthy acquisition processes.

\subsection{Deep Image Deblurring Methods}
The proliferation of large-scale datasets~\cite{GoPro, REDS, DVD} has led to a dramatic upsurge in using deep neural networks for image deblurring~\cite{GoPro, SRN, PSS-NSC, DMPHN, DeblurGAN, DeblurGANv2, RWBI, MPRNet, MIMOUNet, Restormer, NAFNet}. These methods are trained directly to recover the latent images from corresponding observations. Among these methods, ~\cite{GoPro, SRN, PSS-NSC, MIMOUNet} design a multi-scale network for sharp image recovery using a coarse-to-fine strategy. However, the upsample operation in the coarse-to-fine strategy requires expensive runtime. To alleviate this, ~\cite{DMPHN, MPRNet} develop a multi-stage network via stacking multiple sub-networks for image deblurring. Additionally, Kupyn et al.~\cite{DeblurGAN, DeblurGANv2} introduce generative adversarial networks (GANs) into image deblurring for blur removal and details generation. Recently, several researchers have used transformers for image restoration~\cite{IPT, Uformer, Restormer} due to their non-local characteristics. However, existing transformer-based methods only work with single-scale image patches and fail to obtain multi-scale feature representations at the same feature level~\cite{MPViT}. 
\section{Methodology}
\subsection{Motivation}
To motivate our work, we first review the previous blur synthesis methods.

In real-world scenarios, blur is primarily caused by accumulating the signals of the sharp image captured by the camera sensor during the exposure time~\cite{GoPro, RWBI}. It can be modeled as follows:
\begin{equation}
	\label{blur_raw_synthesis}
	I_{b}^{raw}=\frac{1}{T} \int_{t=0}^{T}S(t)dt\simeq \frac{1}{M}\sum_{j=1}^{M}I_{s_{j}}^{raw}
\end{equation}
where $T$ and $S(t)$ denote the exposure time and the signals captured by the camera sensor at time $t$, respectively. $M$ and $I_{s_{j}}^{raw}$ are the number of sampled sharp RAW frames and the $j$-th sharp RAW frame in the recorded videos. The blurry RAW image is then processed by ISP to generate a blurry RGB image, which can be written as:
\begin{equation}
	\label{ISP}
	I_{b}^{rgb}=G(I_{b}^{raw})
\end{equation}
where $G(\cdot)$ and $I_{b}^{rgb}$ denote the ISP and blurry RGB image.

In previous work~\cite{DVD, HIDE, 4KRD}, the researchers generate blurry images in the RGB domain by averaging successive sharp RGB images which treat $G(\cdot)$ as a linear function. However, the ISP pipeline is a nonlinear process which that includes white balance, demosaicing, denoising, color space transformation, and so on~\cite{InvISP, CycleISP, Software}. Consequently, the blur synthesized in the RGB domain does not correspond to the real-world blur. Moreover, Nah et al.~\cite{GoPro, REDS} propose to synthesize the blurry image in the signal space via an estimated camera response function (CRF) and its inverse CRF. However, since the CRF is not trivial to estimate accurate, both approximating it as the gamma function~\cite{GoPro} and measuring it from multi-exposure scene captures~\cite{REDS} are suboptimal. Furthermore, all methods used for synthesizing blur data rely on specific cameras, such as GoPro Hero cameras, impairing the generalization ability of deblurring models for dealing with real-world blurs captured by other hand-held devices~\cite{DADA}.

Based on the above analysis, we can identify two important designs for realistic blur data synthesis. On the one hand, according to Eq.~\ref{blur_raw_synthesis}, the blurs are produced in the RAW domain. On the other hand, Eq.~\ref{ISP} motivates us to design a learnable ISP that can be generalized well to different recording devices. In the following, we describe our realistic blur synthesis pipeline in detail. 
\subsection{Realistic Blur Synthesis Pipeline}
\begin{figure*}[!t]
	\captionsetup[figure]{name={Figure}}
	\centering
	\includegraphics[scale=0.65]{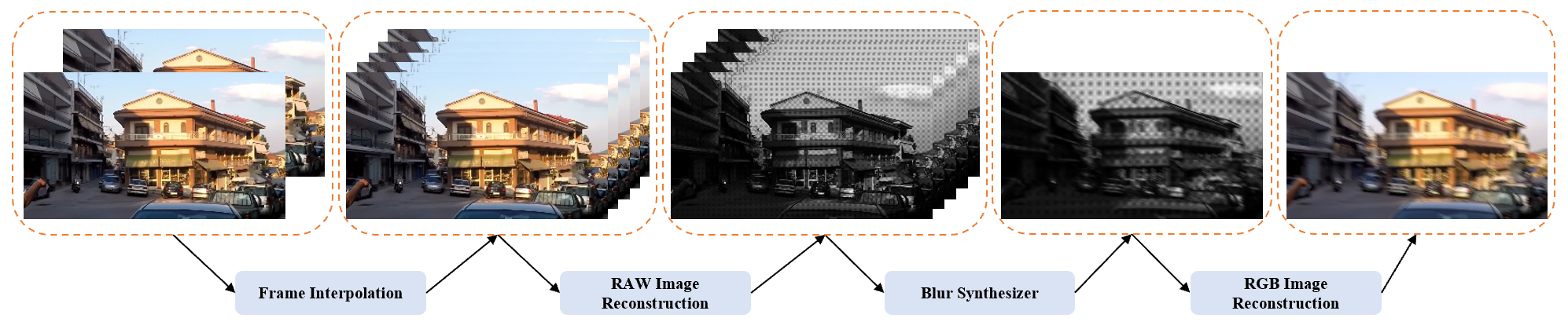}
	\caption{An overview of our pipeline for generating blurred data.}
	\label{pipeline}
\end{figure*} 
The overview of our proposed realistic blur synthesis pipeline can be seen in Figure~\ref{pipeline}. Next, we elaborate the procedures of our pipeline to generate realistic blurry data.

\noindent
{\bf Frame interpolation.} According to~\cite{REDS}, simply averaging successive frames on low-frame-rate videos may cause unnatural blurs with aliasing artifacts. To avoid this, we increase the frame rate using FILM~\cite{FILM} which can synthesize multiple frames between two successive images with large motion.

\noindent
{\bf RAW image reconstruction.} The key feature of our data synthesis pipeline is that blur is performed in the RAW domain which is different from~\cite{DVD, HIDE, 4KRD}. Therefore, we need to convert the RGB sharp frames to RAW frames. To achieve this, we use CycleISP~\cite{CycleISP} which can take RGB image as input and output synthesized RAW image. 

\noindent
{\bf Blur synthesizer.} Based on Eq.~\ref{blur_raw_synthesis}, we average multiple RAW frames to generate blurred ones. Since we want to simulate the blurs caused by varying camera exposures, so the number of averaged frames is random.

\noindent
{\bf RGB image reconstruction.} We generate blurry RGB images with CycleISP~\cite{CycleISP} which can also reconstruct RGB images from realistic RAW data. Due to the color attention uint in CycleISP, the model can be generalized well to different cameras. As a result, our synthesized blurry images can, to some extent, reduce the domain gap between synthesized blurry images and realistic captured blurry images using different hand-held devices~\cite{DADA}. It is worth noting that the corresponding ground truth RGB image is generated via CycleISP taking as input the intermediate sharp RAW frame. For example, if we select 63 successive sharp RAW frames to produce one blurry RAW image, the corresponding sharp RGB image is generated using CycleISP with the input of the 32nd sharp RAW frame.

According to the aforementioned blur synthesis method, we generate 2199 blurry/sharp image pairs. In our experiments, we observe that the deblurring models trained on our proposed training data can generalize well to real-world blurry images, as compared with those trained on existing datasets.  
\subsection{Deblurring Architecture}
\begin{figure*}[!t]
	\captionsetup[figure]{name={Figure}}
	\centering
	\includegraphics[scale=0.6]{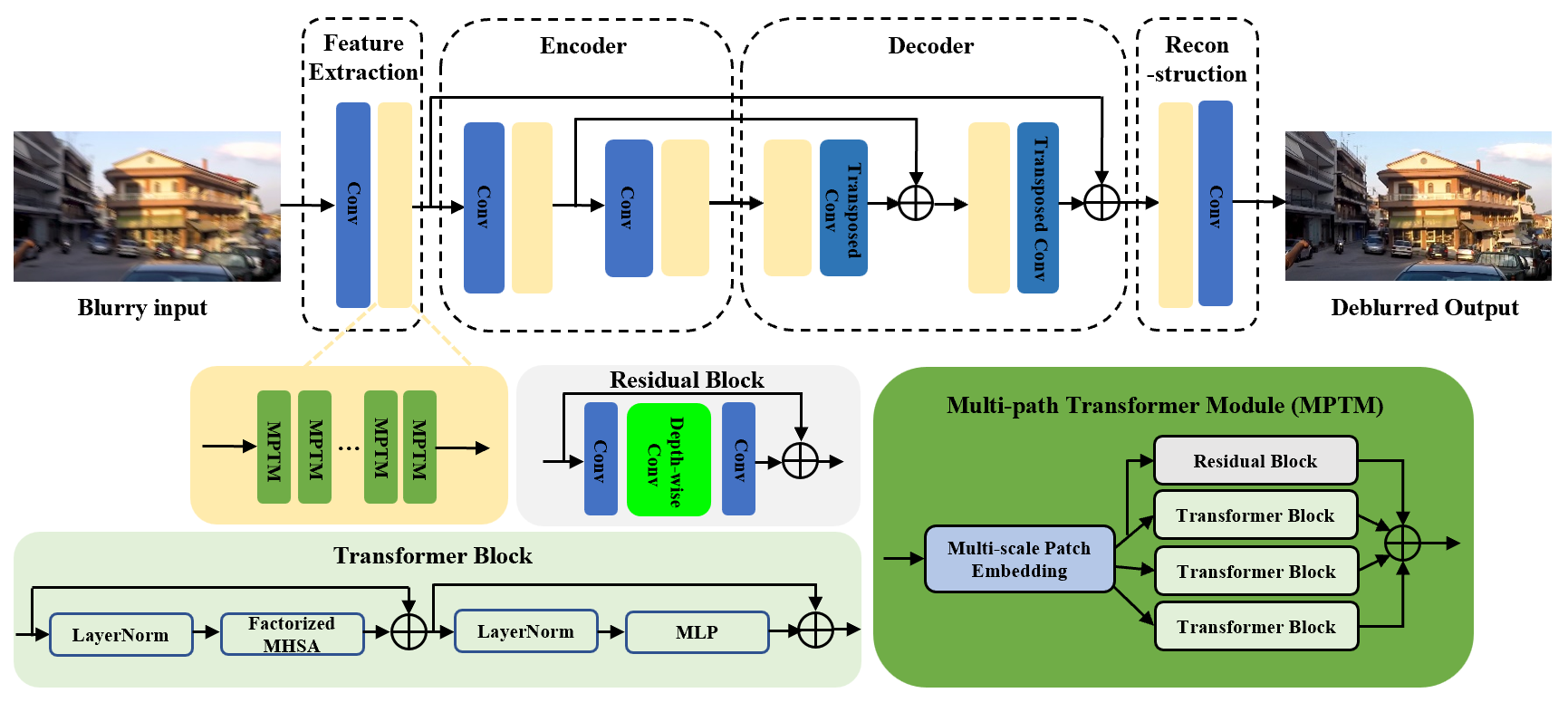}
	\caption{An overview of the MPTUNet architecture and the multi-path transformer module (MPTM). MPTM contains multi-scale patch embedding~\cite{MPViT} and parallel transformer blocks for learning coarse and fine feature representations. The transformer block employs the factorized multi-head self-attention (MHSA)~\cite{MHSA} for alleviating the computational burden and capturing the long-range dependencies. The residual block aims to capture local dependencies and consists of 1$\times$1 convolution, 3$\times$3 depthwise convolution, and 1$\times$1 convolution and skip connection.}
	\label{architecture}
\end{figure*}
Figure~\ref{architecture} shows the architecture of MPTUNet. Based on symmetric encoder-decoder architecture, MPTUNet enables information aggregation at different levels by utilizing skip connections.

\noindent
{\bf Feature extraction.} Given a blurry input $I_{b}^{rgb} \in \mathbb{R}^{H\times W\times 3}$, we first extract the shallow features $F_{s_{0}} \in \mathbb{R}^{H\times W\times C}$ of the blurry input which can be expressed as 
\begin{equation}
	F_{s_{0}} = N_{SF}(I_{b}^{rgb}),
\end{equation}
where  $H\times W$ represents the spatial size and $C$ represents the number of features. The shallow feature extraction module $N_{SF}(\cdot)$ contains one 5$\times$5 convolution followed by one multi-path transformer module. 

\noindent
{\bf Encoder and decoder.} In the encoder part, we employ $K$ encoder levels to obtain multi-scale feature representations. Each level consists of one 5$\times$5 convolution with stride 2 and several multi-path transformer modules. The strided convolution aims to downsample the feature maps to half their spatial size and double their channels. In general, the encoding process can be described as 
\begin{equation}
F_{e_{i}}= \left\{
\begin{array}{l}
	N_{E_{i}}(F_{s_{0}}), i=1, \\
	N_{E_{i}}(F_{e_{i-1}}), i=2,...,K,\\
\end{array}
\right.
\end{equation}
where $N_{E_{i}}(\cdot)$ denotes the $i$-th encoder level and $F_{e_{i}} \in \mathbb{R}^{\frac{H}{2^{i}}\times \frac{W}{2^{i}}\times 2^{i}C}$ denote encoding features. For progressively obtaining high-resolution feature representations, we use $K$ levels in the decoder part which is symmetric to the encoder. Each level consists of a stack of multi-path transformer modules and one 5$\times$5 transposed convolution. By using transposed convolution. the features are doubled in spatial size and halved in number. It is worth noting that the upsampling features in the decoder are aggregated with the corresponding encoder features via skip connections. We formulate the decoding process as follows,
\begin{equation}
F_{d_{i}}= \left\{
\begin{array}{l}
	N_{D_{i}}(F_{e_{i}}), i=K, \\
	N_{D_{i}}(F_{d_{i+1}}+F_{e_{i}}), i=K-1,...2,1,\\
\end{array}
\right.
\end{equation}
where $N_{D_{i}}$ is the $i$-th decoder level and $F_{d_{i}}\in \mathbb{R}^{\frac{H}{2^{i-1}}\times \frac{W}{2^{i-1}}\times 2^{i-1}C}$ are upsample features. Following the decoder, the enriched features are sent to the next reconstruction module for recovery.

\noindent
{\bf Reconstruction.} Take as input the decoded features $F_{d_{1}}$ and shallow features $F_{s_{0}}$, the reconstruction module directly generates the deblurred output $\hat{I}_{s}^{rgb}$ as
\begin{equation}
	\hat{I}_{s}^{rgb} = N_{REC}(F_{d_{1}} + F_{s_{0}}),
\end{equation}
where $N_{REC}(\cdot)$ represents the reconstruction module, which comprises one multi-path transformer module and 5$\times$5 convolution layer. Except for the last convolution layer, all convolution and transposed convolution layers are followed by ReLU activation function.

\noindent
{\bf Multi-path transformer module.} With coarse and fine feature representations providing precise spatial and contextual information, respectively, multi-scale features are essential to various image restoration tasks~\cite{MIRNet, MIRNetv2}. Hence, we hope that the network architecture will be able to obtain the multi-scale feature representations at the same feature level. Specifically, we introduce the multi-path transformer module (MPTM)~\cite{MPViT} as our main building block. As shown in Figure~\ref{architecture}, MPTM consists of multi-scale patch embedding~\cite{MPViT} followed by one residual block and three parallel transformer blocks. Parallel transformer blocks are used for learning multi-scale non-local features, while the residual block enhances the local features. As compared to ~\cite{MPViT}, we remove the Batch Normalization~\cite{BN} that degrades the performance of restoration models~\cite{ESRGAN, GoPro, SRN}. We also use summation operation to aggregate the multi-path features instead of concatenation operation. More details of MPTM can be found in the supplementary file.

\section{Experiment}
\subsection{Datasets and Implementation}
\noindent
{\bf Training datasets.} We adopt the widely used GoPro (gamma corrected version)~\cite{GoPro}, REDS~\cite{REDS}, and our synthetic dataset in the experiments. The GoPro dataset contains 2103 training images with dynamic motion blur. Similarly, the REDS dataset consists of 24000 training images that cover a wide variety of scenes. To ease our training burden, we select 2300 training images from REDS.

\noindent
{\bf Testing datasets.} We collect three different {\bfseries real-world} datasets from multiple sources. There is no overlap between any of these datasets and the training datasets. A brief introduction is given below.
\begin{itemize}
	\item \emph{BSD}~\cite{BSD} contains 9000 aligned blurry/sharp image pairs captured with a beam splitter system. According to the exposure time setup in~\cite{BSD}, we divide them into three partitions (denoted as \emph{1ms-8ms}, \emph{2ms-16ms}, and \emph{3ms-24ms}, respectively), with each partition containing 3000 testing images.
	\item \emph{Lai}~\cite{Lai} contains 100 real blurred images captured from the wild. They are low-quality and captured using different cameras and settings in real-world scenarios.
	\item \emph{RWBI}~\cite{RWBI} contains 3112 real blurry images taken with various devices, including a Huawei P30 Pro, a Samsung S9 Plus, an iPhone XS, and a GoPro Hero Camera.
\end{itemize}
In our experiments, we only provide visual comparisons of \emph{Lai} and \emph{RWBI} datasets since neither have corresponding ground truth images.

\noindent
{\bf Implementation details.} Follow the setting in ~\cite{MIMOUNet}, we use Adam optimizer~\cite{Adam} to optimize the parameters of MPTUNet with ${L}_{1}$ loss function. The learning rate is fixed to 1e-4. In the training process, the network is trained on 256$\times$256 image patches, randomly cropped from training images. The batch size is set to 4 for 500 epochs. For data augmentation, we perform horizontal and vertical flips. For MPTUNet, the level $K$ of encoder and decoder is set to 2, respectively. And the first level contains one MPTM and the next level contains three MPTMs. Our model is implemented on a NVIDIA GeForce RTX 3090 GPU in Pytorch. The source codes and model will be available on the authors' website upon acceptance. More details and experimental results are presented in the supplementary file. 
\subsection{Comparisons of Different Training Datasets}
\begin{table*}
	\centering
	\begin{tabular}{c|c|cc|cc|cc||cc} 
		\hline
		\multirow{2}{*}{Model}     & \multirow{2}{*}{Training Dataset} & \multicolumn{2}{c|}{\emph{1ms-8ms}} & \multicolumn{2}{c|}{\emph{2ms-16ms}} & \multicolumn{2}{c||}{\emph{3ms-24ms}}   & \multicolumn{2}{c}{Average} \\
		&                                   & PSNR$\uparrow$  & SSIM$\uparrow$          & PSNR$\uparrow$  & SSIM$\uparrow$           & PSNR$\uparrow$  & SSIM$\uparrow$     & PSNR$\uparrow$  & SSIM$\uparrow$     \\ 
		\hline
		\hline
		\multirow{3}{*}{MIMO-UNet} & GoPro  & 25.03 & 0.784                        & 25.27 & 0.788                        & 25.82 & 0.816  & 25.38 & 0.796                    \\
		& REDS   & \bfseries29.14 & 0.868             & 26.91 & 0.820                        & 27.06 & 0.838  & 27.70 & 0.842                  \\
		& Ours   & 28.67 & \bfseries0.874             & \bfseries27.49 & \bfseries0.854  & \bfseries27.54 & \bfseries0.862  & \bfseries27.90 & \bfseries0.863 \\ 
		\hline
		\hline
		\multirow{3}{*}{MIRNet-v2} & GoPro  & 29.77 & 0.886                        & 28.35 & 0.858                    & 28.22 & 0.869   & 28.78 & 0.871                  \\
		& REDS   & 29.28 & 0.872                        & 27.38 & 0.830                    & 27.60 & 0.851   & 28.09 & 0.851                  \\
		& Ours   & \bfseries30.30 & \bfseries0.907  & \bfseries29.02 & \bfseries0.882  & \bfseries28.43 & \bfseries0.878 & \bfseries29.25 & \bfseries0.889 \\ 
		\hline
		\hline
		\multirow{3}{*}{NAFNet}    & GoPro  & 29.71 & 0.882                        & 28.82 & 0.867                    & \bfseries29.07 & 0.889            & 29.20 & 0.879         \\
		& REDS   & 30.16 & 0.893                        & 28.42 & 0.865                    & 28.70 & 0.882 & 29.09 & 0.880                     \\
		& Ours   & \bfseries30.67 & \bfseries0.912  & \bfseries29.49 & \bfseries0.895 & 28.79 & \bfseries0.891  & \bfseries29.65 & \bfseries0.899 \\ 
		\hline
		\hline
		\multirow{3}{*}{MPTUNet}   & GoPro  & 29.87 & 0.895                        & 28.83 & 0.872                    & 28.48 & 0.881         & 29.06 & 0.882            \\
		& REDS   & 30.12 & 0.895                        & 28.56 & 0.868                    & 28.40 & 0.875         & 29.03 & 0.879                     \\
		& Ours   & \bfseries30.66 & \bfseries0.914  & \bfseries29.41 & \bfseries0.893 & \bfseries28.71 & \bfseries0.889  & \bfseries29.59 & \bfseries0.899 \\
		\hline
	\end{tabular}
	\caption{Quantitative comparisons among different training datasets. All the deblurring models are trained in the same experimental setting and evaluated on the \emph{BSD} dataset by calculating the PSNR and SSIM.}
	\label{TC}
\end{table*}
\begin{figure*}[h]
	\captionsetup[figure]{name={Figure}}
	\centering
	\subfigure[Blurry input]{
		\begin{minipage}[b]{.18\linewidth}
			\centering
			\includegraphics[width=\textwidth]{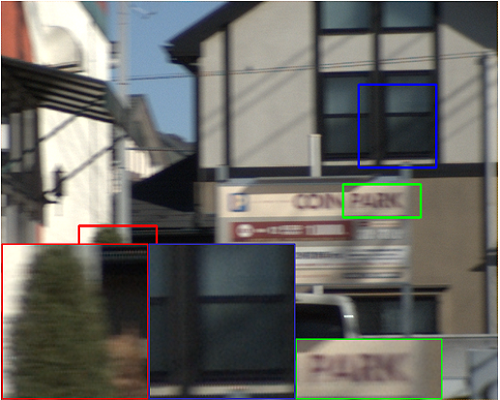}\\
			\includegraphics[width=\textwidth]{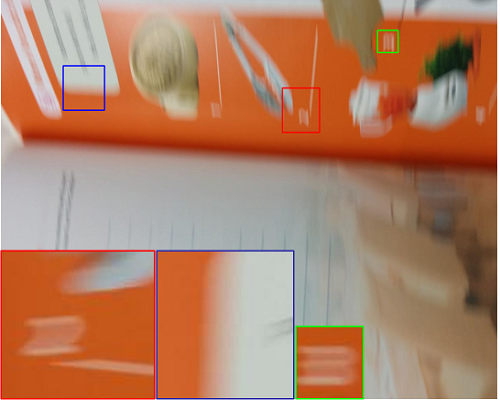}\\
			\includegraphics[width=\textwidth]{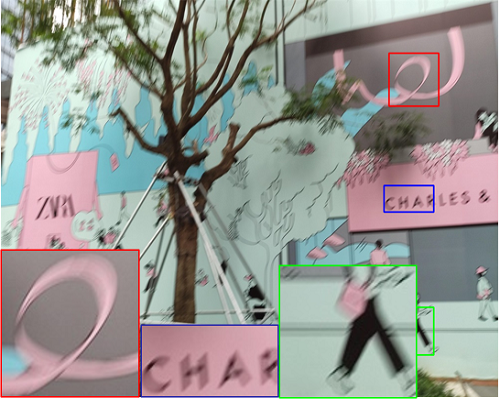}
		\end{minipage}
		\label{TCV:Blurry}
	}
	\subfigure[MIRNet-v2 (GoPro)]{
		\begin{minipage}[b]{.18\linewidth}
			\centering
			\includegraphics[width=\textwidth]{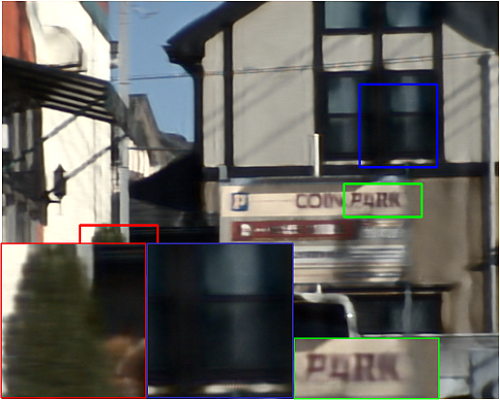}\\
			\includegraphics[width=\textwidth]{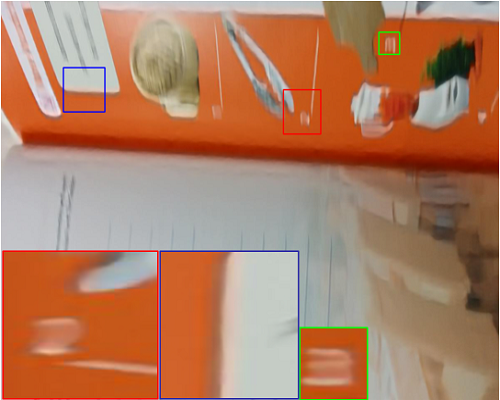}\\
			\includegraphics[width=\textwidth]{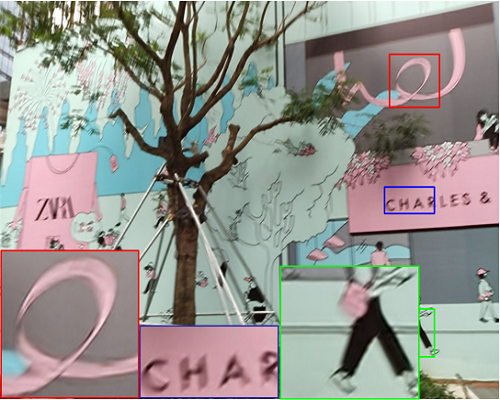}
		\end{minipage}
		\label{TCV:GoPro}
	}
	\subfigure[MIRNet-v2 (REDS)]{
		\begin{minipage}[b]{.18\linewidth}
			\centering
			\includegraphics[width=\textwidth]{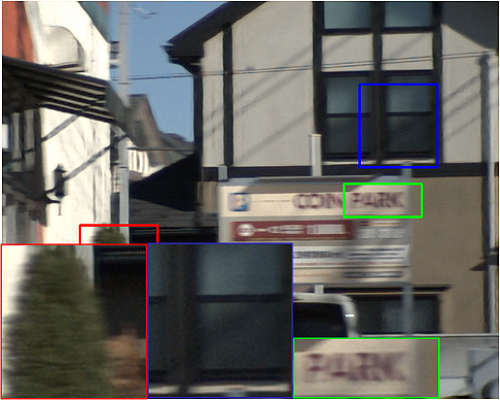}\\
			\includegraphics[width=\textwidth]{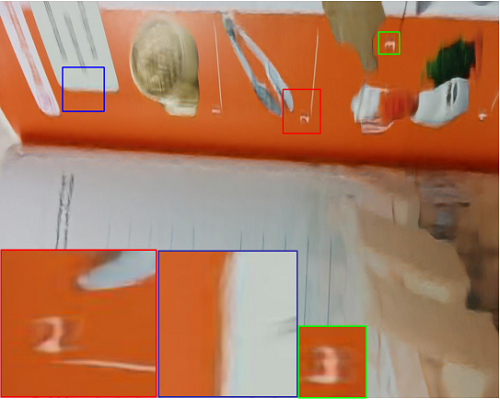}\\
			\includegraphics[width=\textwidth]{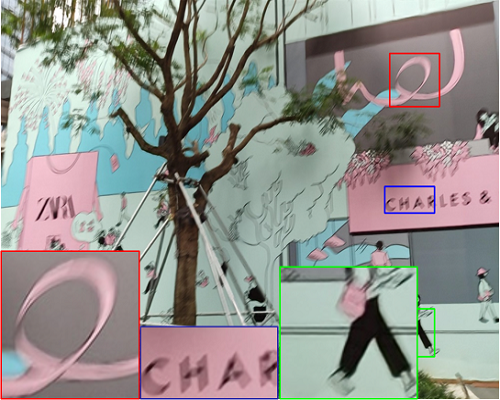}
		\end{minipage}
		\label{TCV:REDS}
	}
	\subfigure[MIRNet-v2 (Ours)]{
		\begin{minipage}[b]{.18\linewidth}
			\centering
			\includegraphics[width=\textwidth]{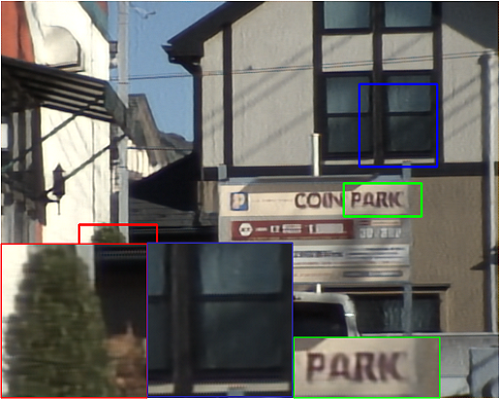}\\
			\includegraphics[width=\textwidth]{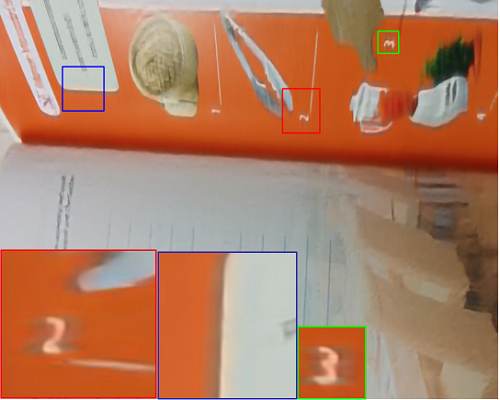}\\
			\includegraphics[width=\textwidth]{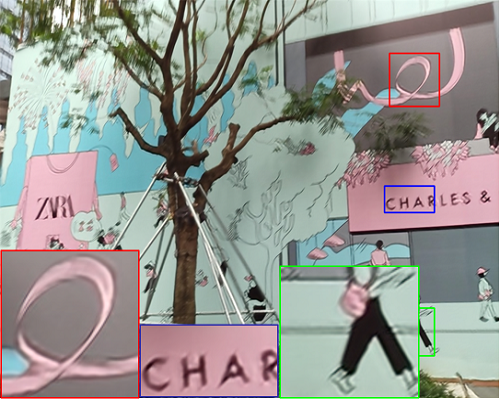}
		\end{minipage}
		\label{TCV:Ours}	
	}
	\subfigure[Ground truth]{
		\begin{minipage}[b]{.18\linewidth}
			\centering
			\includegraphics[width=\textwidth]{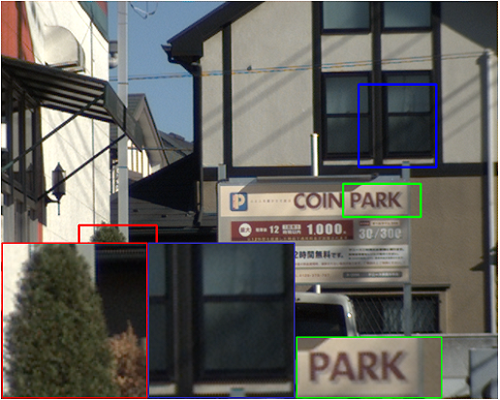}\\
			\includegraphics[width=\textwidth]{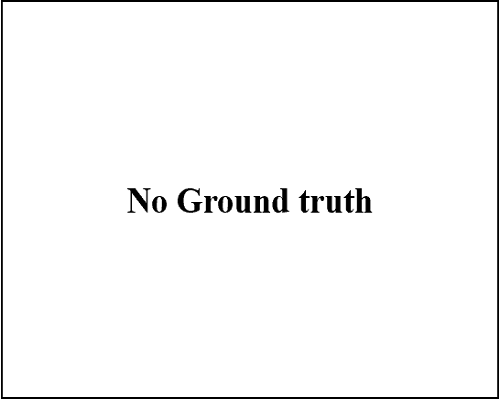}\\
			\includegraphics[width=\textwidth]{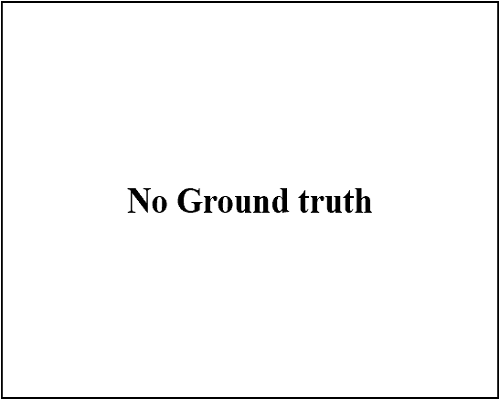}
		\end{minipage}
		\label{TCV:GT}
	}
	\caption{Qualitative comparisons among different training datasets. From top to bottom, the images are from \emph{BSD}, \emph{Lai} and \emph{RWBI}. The MIRNet-v2 model trained on our synthetic data greatly improves the visual quality and sharpness of the deblurred images.}
	\label{TCV}
\end{figure*} 
To assess the effectiveness of our proposed blur synthesis pipeline, we compare our synthetic data with the representative deblurring datasets such as GoPro~\cite{GoPro} and REDS~\cite{REDS}. We train four restoration models (MIMO-UNet~\cite{MIMOUNet}, MIRNet-v2~\cite{MIRNetv2}, NAFNet~\cite{NAFNet} and our proposed MPTUNet) from scratch using the same experimental settings as mentioned above. After training, we evaluate competing models on the \emph{BSD} dataset in terms of PSNR and SSIM~\cite{SSIM}. As shown in Table~\ref{TC}, we can observe that our synthetic training data can help improve the generalization capability of the deblurring models. For example, MIRNet-v2 trained on our synthetic data improves its performance by 0.47 dB and 1.16 dB over its trained on GoPro and REDS, respectively.

Qualitative results are presented in Figure~\ref{TCV}. Here we take MIRNet-v2 as an example. When trained on GoPro or REDS, MIRNet-v2 either generates images with severe artifacts or fails to remove the real blur. By contrast, MIRNet-v2 trained with our synthetic data can effectively remove the blur and produce clean results (see Figure~\ref{TCV:Ours}). This validates the advantage of our synthetic training data that can help improve the generalization ability of deblurring models to handle real-world blur.

\subsection{Comparisons with State-of-the-art Methods}
\begin{table*}[!t]
	\centering
	\begin{tabular}{c|c|cc|cc|cc||cc} 
		\hline
		\multirow{2}{*}{Model}     & \multirow{2}{*}{\# Param} & \multicolumn{2}{c|}{\emph{1ms-8ms}} & \multicolumn{2}{c|}{\emph{2ms-16ms}} & \multicolumn{2}{c||}{\emph{3ms-24ms}}   & \multicolumn{2}{c}{Average} \\
		&                                   & PSNR$\uparrow$  & SSIM$\uparrow$          & PSNR$\uparrow$  & SSIM$\uparrow$           & PSNR$\uparrow$  & SSIM$\uparrow$     & PSNR$\uparrow$  & SSIM$\uparrow$     \\ 
		\hline
		\hline
		MIMO-UNet & 6.81M & 28.67 & 0.874  & 27.49 & 0.854  & 27.54 & 0.862  & 27.90 & 0.863 \\ 
		MIRNet-v2 & 0.94M & 30.30 & 0.907  & 29.02 & 0.882  & 28.43 & 0.878  & 29.25 & 0.889 \\ 
		NAFNet    & 17.11M & 30.67 & 0.912  & 29.49 & \bfseries0.895  & 28.79 & \bfseries0.891  & 29.65 & 0.899 \\ 
		\hline
		MPTUNet   & 3.85M & 30.66 & \bfseries0.914  & 29.41 & 0.893  & 28.71 & 0.889  & 29.59 & 0.899 \\
		MPTUNet+  & 5.7M & \bfseries30.95 & \bfseries0.914  & \bfseries29.65 & \bfseries0.895 & \bfseries28.89 & 0.890  & \bfseries29.83 & \bfseries0.900 \\
		\hline
	\end{tabular}
	\caption{Quantitative comparisons of different deblurring models. All the deblurring models are trained on our synthetic data and evaluated on the \emph{BSD} dataset by calculating the PSNR and SSIM.}
	\label{MC}
\end{table*}
\begin{figure*}[!t]
	\captionsetup[figure]{name={Figure}}
	\centering
	\subfigure[Blurry input]{
		\begin{minipage}[b]{.15\linewidth}
			\centering
			\includegraphics[width=\textwidth]{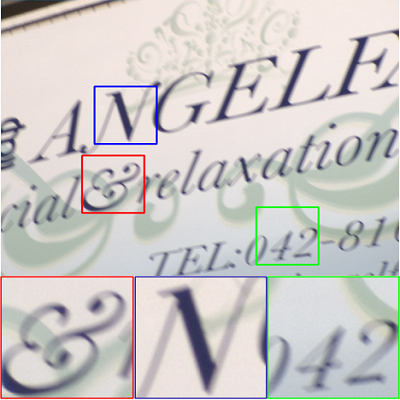}\\
			\includegraphics[width=\textwidth]{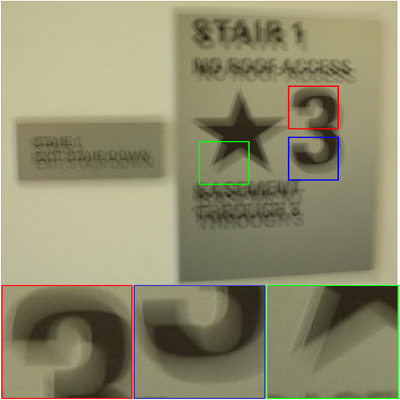}\\
			\includegraphics[width=\textwidth]{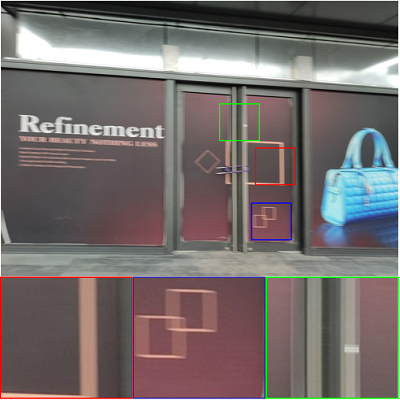}
		\end{minipage}
		\label{MCV:Blurry}
	}
	\subfigure[MIMO-UNet]{
		\begin{minipage}[b]{.15\linewidth}
			\centering
			\includegraphics[width=\textwidth]{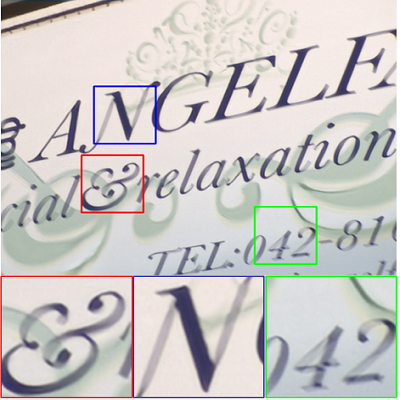}\\
			\includegraphics[width=\textwidth]{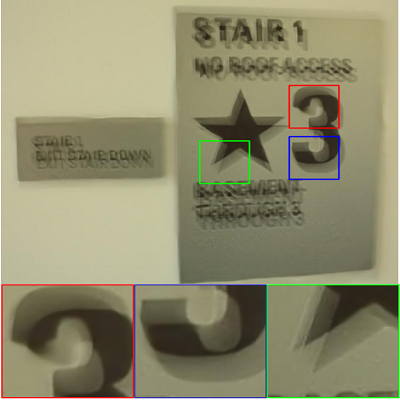}\\
			\includegraphics[width=\textwidth]{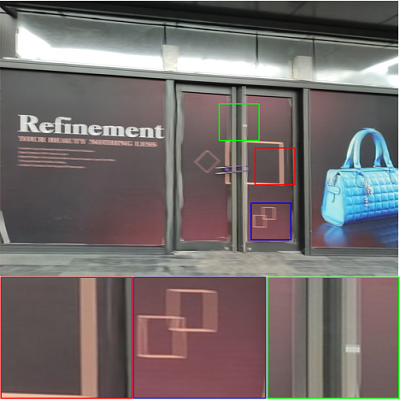}
		\end{minipage}
		\label{MCV:MIMOUNet}
	}
	\subfigure[MIRNet-v2]{
		\begin{minipage}[b]{.15\linewidth}
			\centering
			\includegraphics[width=\textwidth]{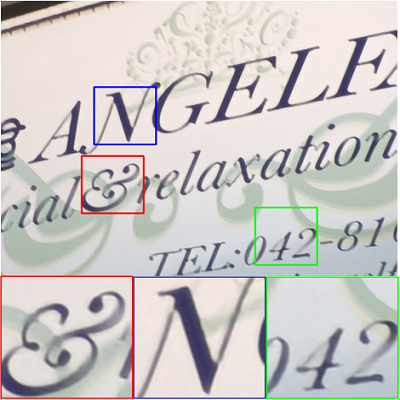}\\
			\includegraphics[width=\textwidth]{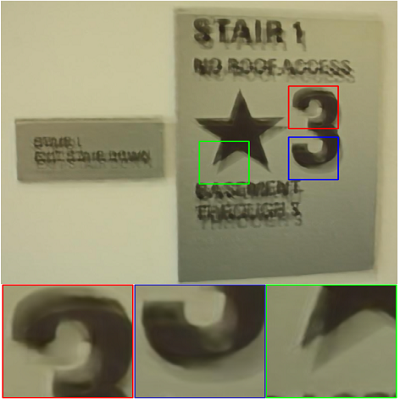}\\
			\includegraphics[width=\textwidth]{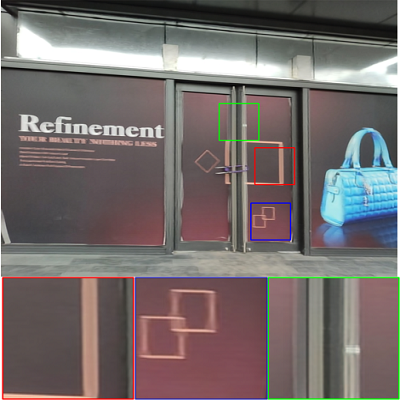}
		\end{minipage}
		\label{MCV:MIRNetv2}
	}
	\subfigure[NAFNet]{
		\begin{minipage}[b]{.15\linewidth}
			\centering
			\includegraphics[width=\textwidth]{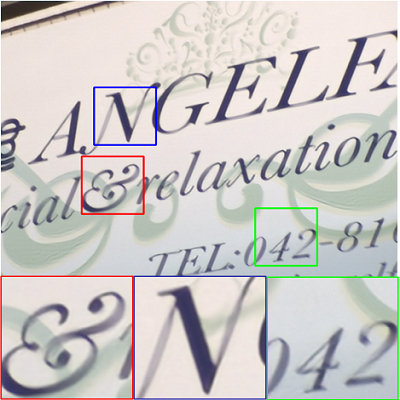}\\
			\includegraphics[width=\textwidth]{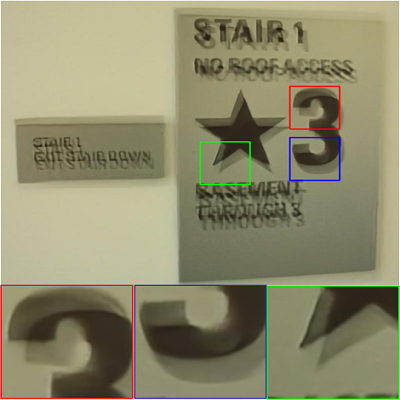}\\
			\includegraphics[width=\textwidth]{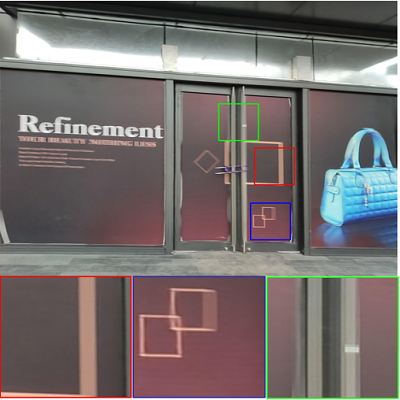}
		\end{minipage}
		\label{MCV:NAFNet}	
	}
	\subfigure[MPTUNet]{
	\begin{minipage}[b]{.15\linewidth}
		\centering
		\includegraphics[width=\textwidth]{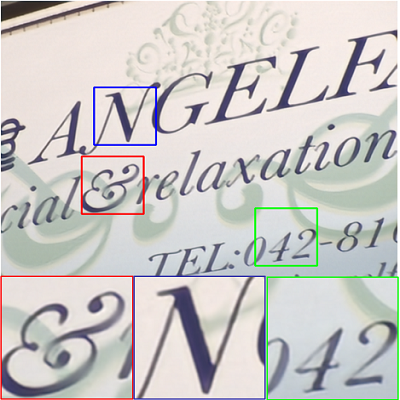}\\
		\includegraphics[width=\textwidth]{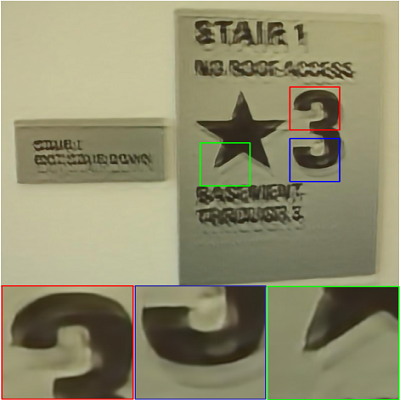}\\
		\includegraphics[width=\textwidth]{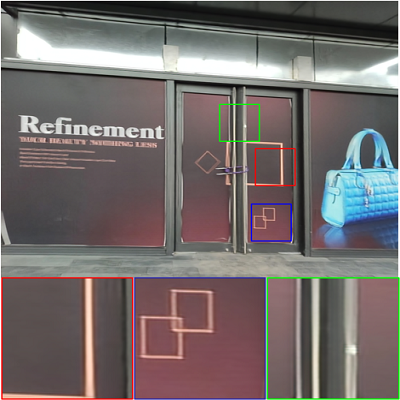}
	\end{minipage}
	\label{MCV:MPTUNet}
}
	\subfigure[Ground truth]{
		\begin{minipage}[b]{.15\linewidth}
			\centering
			\includegraphics[width=\textwidth]{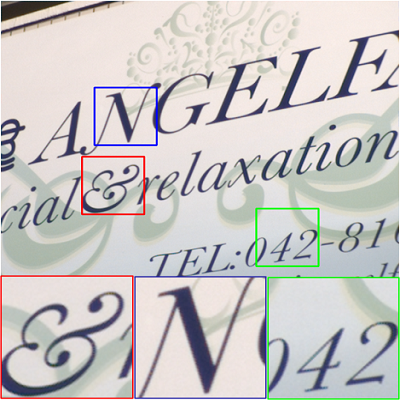}\\
			\includegraphics[width=\textwidth]{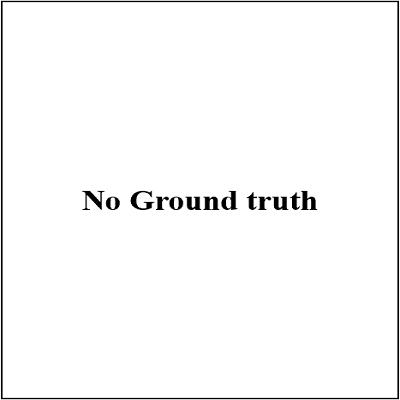}\\
			\includegraphics[width=\textwidth]{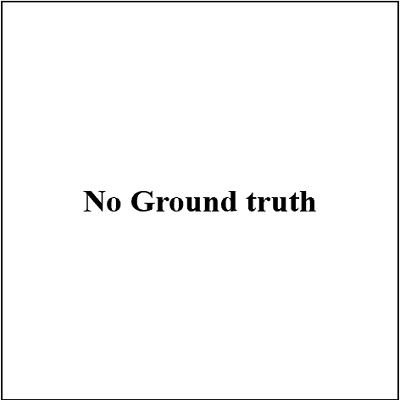}
		\end{minipage}
		\label{MCV:GT}
	}
	\caption{Qualitative comparisons with different deblurring models. The first row is from the \emph{BSD} dataset and the last two rows are from the \emph{Lai} dataset and \emph{RWBI} dataset, respectively.}
	\label{MCV}
\end{figure*} 
We compare proposed MPTUNet with the state-of-the-art restoration models: MIMO-UNet~\cite{MIMOUNet}, MIRNet-v2~\cite{MIRNetv2}, and NAFNet~\cite{NAFNet}. 
We employ a two-stage strategy~\cite{DMPHN}, which we call MPTUNet+, to improve the performance of MPTUNet. MPTUNet+ is the stacked version of MPTUNet-tiny and MPTUNet, where MPTUNet-tiny only employs a single MPTM at each level of the encoder/decoder part. Table~\ref{MC} displays PSNR and SSIM values for deblurring on the \emph{BSD} dataset. From the perspective of quantitative analysis, MPTUNet+ has an advantage over the state-of-the-art restoration methods, as well as 11.41M fewer parameters than NAFNet. In addition, MPTUNet also achieves comparable performance in comparison with other methods. 

Figure~\ref{MCV} shows the deblurred results on different real-world datasets for qualitative comparisons. Compared with other approaches, MPTUNet could produce sharper text and structure information.
\section{Discussion and analysis}
\noindent
{\bf Our blur synthesis method vs. ~\cite{DVD, HIDE, 4KRD}.} Unlike previous blur synthesis methods~\cite{DVD, HIDE, 4KRD}, we propose synthesizing blur in the RAW domain rather than the RGB domain. By directly operating on the raw sensor data, our method can produce more realistic blurred images that are consistent with the blur generation process. To validate our conjecture, we use the same blur generation methods as~\cite{DVD, HIDE, 4KRD} for synthesizing training data (denoted as \emph{Our-RGB}) in the RGB domain. Note that the contents of \emph{Our-RGB} are similar to those of our proposed training data (denoted as \emph{Our-RAW}).

The quantitative comparisons between \emph{Our-RGB} and \emph{Our-RAW} are shown in Table~\ref{pipline_compare}. We can observe that all the deblurring models trained using \emph{Our-RAW} exhibit a greater generalization capability to handle real-world blur than those trained using \emph{Our-RGB}. As an example, \emph{Our-RAW} can improve the PSNR of \emph{Our-RGB} trained MIMO-UNet by 0.23 dB, and improvements can also be noticed for MIRNet-v2, NAFNet, and MPTUNet, where the PSNR is increased by 0.88 dB, 0.63 dB, and 0.99 dB, respectively. In summary, all of the above demonstrate that the blur synthesized in the RAW domain mimics real-world blur more closely than that synthesized in the RGB domain. 

\begin{figure}[!t]
	\captionsetup[figure]{name={Figure}}
	\centering
	\subfigure[Input]{
		\label{PV:Blurry input}
		\includegraphics[scale=0.5]{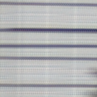}
	}
	\subfigure[Path 1]{
		\label{PV:path1}
		\includegraphics[scale=0.5]{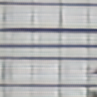}
	}
	\subfigure[Path 2]{
		\label{PV:path2}
		\includegraphics[scale=0.5]{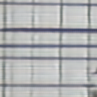}
	}
	\subfigure[Path 3]{
		\label{PV:path3}
		\includegraphics[scale=0.5]{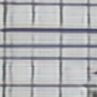}
	}
	\caption{Effectiveness of the number of paths for real-world image deblurring. (a) is the blurry input. (b)-(d) denote the deblurred results from path 1, 2, and 3, respectively. By increasing the number of paths, the MPTUNet can produce sharper results.}
	\label{PV}
\end{figure}

\begin{table*}[!t]
	\centering
		\begin{tabular}{c|c|cc|cc|cc||cc} 
			\hline
			\multirow{2}{*}{Model}     & \multirow{2}{*}{Training Dataset} & \multicolumn{2}{c|}{\emph{1ms-8ms}} & \multicolumn{2}{c|}{\emph{2ms-16ms}} & \multicolumn{2}{c||}{\emph{3ms-24ms}} & \multicolumn{2}{c}{Average}  \\
			&                                   & PSNR$\uparrow$  & SSIM$\uparrow$          & PSNR$\uparrow$  & SSIM$\uparrow$           & PSNR$\uparrow$  & SSIM$\uparrow$   & PSNR$\uparrow$  & SSIM$\uparrow$        \\ 
			\hline
			\hline
			\multirow{2}{*}{MIMO-UNet} & \emph{Our-RGB}   & 29.13 & 0.873                        & 26.81 & 0.825                     & 27.06 & 0.842    & 27.67 & 0.847                \\
			& \emph{Our-RAW}   & 28.67 & 0.874                        & 27.49 & 0.854                     & 27.54 & 0.862    & 27.90 & 0.863                \\
			Improvement                &            & \bfseries-0.46 & \bfseries+0.001     & \bfseries+0.68 & \bfseries+0.029  & \bfseries+0.48 & \bfseries+0.02  & \bfseries+0.23 & \bfseries+0.016\\ 
			\hline
			\multirow{2}{*}{MIRNet-v2} & \emph{Our-RGB}   & 29.47 & 0.890                        & 27.88 & 0.860                     & 27.74 & 0.867    & 28.37 & 0.872               \\
			& \emph{Our-RAW}   & 30.30 & 0.907                        & 29.02 & 0.882                     & 28.43 & 0.878    & 29.25 & 0.889                \\
			Improvement                &            & \bfseries+0.83 & \bfseries+0.017      & \bfseries+1.14 & \bfseries+0.022  & \bfseries+0.69 & \bfseries+0.011 &\bfseries+0.88 & \bfseries+0.017 \\ 
			\hline
			\multirow{2}{*}{NAFNet}    & \emph{Our-RGB}   & 30.14 & 0.902                        & 28.68 & 0.882                     & 28.23 & 0.884    & 29.02 & 0.890                 \\
			& \emph{Our-RAW}   & 30.67 & 0.912                        & 29.49 & 0.895                     & 28.79 & 0.891    & 29.65 & 0.899                 \\
			Improvement                &            & \bfseries+0.53 & \bfseries+0.01      & \bfseries+0.81 & \bfseries+0.013  & \bfseries+0.56 & \bfseries+0.004 &\bfseries+0.63 & \bfseries+0.009   \\ 
			\hline
			\multirow{2}{*}{MPTUNet}   & \emph{Our-RGB}   & 29.72 & 0.907                        & 28.20 & 0.882                     & 27.87 & 0.884    & 28.60 & 0.891                  \\
			& \emph{Our-RAW}   & 30.66 & 0.914                        & 29.41 & 0.893                     & 28.71 & 0.889    & 29.59 & 0.899                  \\
			Improvement                &           & \bfseries+0.94  &  \bfseries+0.007     &  \bfseries+0.21 & \bfseries+0.011 & \bfseries+0.84 & \bfseries+0.005 & \bfseries+0.99 & \bfseries+0.008   \\
			\hline
		\end{tabular}
	\caption{Quantitative comparisons between \emph{Our-RGB} and \emph{Our-RAW}. All the deblurring models are trained in the same experimental settings and evaluated on the \emph{BSD} dataset.}
	\label{pipline_compare}
\end{table*}
\noindent
{\bf Number of the paths.} In Table~\ref{experiment_paths} and Figure~\ref{PV}, we examine how the number of paths affects the real-world image deblurring results. It is noteworthy that increasing the number of paths results in better-deblurred images, therefore supporting the use of multi-scale feature representations exploration for image deblurring. In our experiment, we choose three paths based on the trade-off between deblurring accuracy and model capacity. 

\begin{table}[!t]
	\centering
	\begin{tabular}{c|ccc} 
		\hline
		Path & PSNR & SSIM  & \# Param   \\ 
		\hline
		\hline
		1    & 29.42     & 0.894 & 1.78M        \\
		2    & 29.52     & 0.896 & 2.81M        \\
		3    & \bfseries29.59     & \bfseries0.899 & 3.85M        \\
		\hline
	\end{tabular}
	\caption{Experiments to determine the number of paths. The average PSNR and SSIM values are computed on the real-world \emph{BSD} dataset.}
	\label{experiment_paths}
\end{table}
\noindent
{\bf Feature aggregation.} To aggregate the features from multiple paths, we directly employ the summation operation rather than concatenation operation followed by 1$\times$1 convolution layer. The results are shown in Table~\ref{experiment_fusion}. We can observe that MPTUNet with the summation operation has fewer parameters and achieves better performance than it with the concatenation operation.

\begin{figure}[!t]
	\captionsetup[figure]{name={Figure}}
	\centering
	\subfigure[Blurry input]{
		\begin{minipage}[b]{.45\linewidth}
			\centering
			\includegraphics[width=\textwidth]{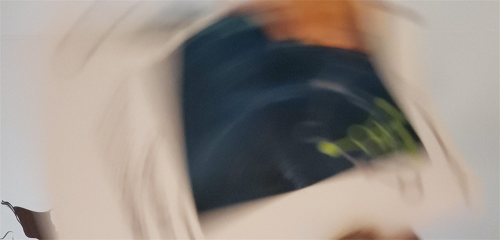}\\
			\includegraphics[width=\textwidth]{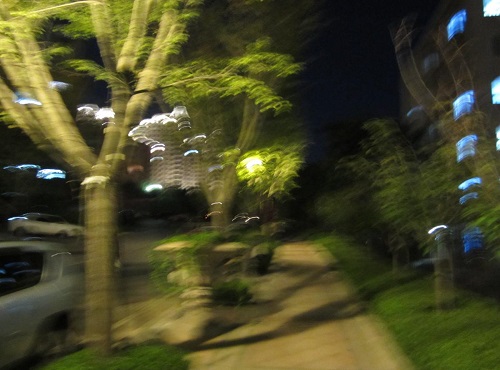}
		\end{minipage}
		\label{Blurry input}
	}
	\subfigure[Deblurred output]{
		\begin{minipage}[b]{.45\linewidth}
			\centering
			\includegraphics[width=\textwidth]{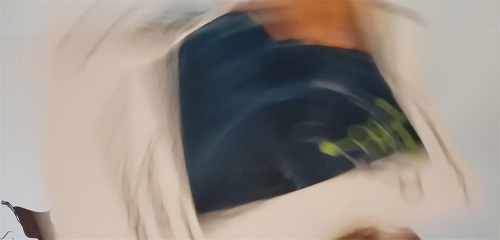}\\
			\includegraphics[width=\textwidth]{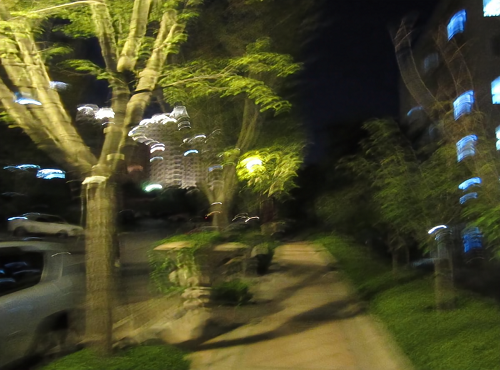}
		\end{minipage}
		\label{Deblurred}
	}
	\caption{Limitations: severe blur caused by large motions (the top row) and night blurry image with saturated regions (the bottom row).}
	\label{Limitation}
\end{figure}

\begin{table}[!t]
	\centering
	\begin{tabular}{c|ccc} 
		\hline
		Aggregation & PSNR & SSIM  & \# Param  \\ 
		\hline
		\hline
		concatenation    & 29.53     & 0.895 & 4.28M        \\
		summation    & \bfseries29.59     & \bfseries0.899 & \bfseries3.85M        \\
		\hline
	\end{tabular}
	\caption{Comparison of concatenation (w/ 1$\times$1 convolution layer that reduces the channels) and summation. The average PSNR and SSIM values are computed on the \emph{BSD} dataset.}
	\label{experiment_fusion}
\end{table}
\noindent
{\bf Limitations and future works.} Figure~\ref{Limitation} illustrates the limitations of our proposed MPTUNet. It can be seen that MPTUNet fails to remove the severe blurs caused by large motions and blurs in saturated areas. To overcome these problems, we can directly expand our training data to cover a greater variety of blurs and complex scenarios. Furthermore, it is necessary to address how to synthesize more realistic degraded images that can cover a broader range of degradations (e.g., noise, compression artifacts, low-night environments) in future works, which are crucial for practical applications.    

\section{Conclusion}
In this paper, we present a novel blur synthesis pipeline that mimics realistic blur generation. Specifically, we have highlighted two key designs, namely, blur synthesis in the RAW domain and a learnable ISP for RGB image reconstruction that is robust to different imaging devices. The synthetic data generated by the proposed blur synthesis method can improve the performance of existing deblurring algorithms for real blur removal. We also propose an effective deblurring model, MPTUNet, that possesses both local and non-local modeling ability. Experimental results demonstrate the capability of MPTUNet to restore real-world blurry images and to perform favorably against state-of-the-art methods.

\bibliography{aaai23}
\end{document}